# A Hybrid MLP-SVM Model for Classification using Spatial-Spectral Features on Hyper-Spectral Images

Ginni Garg*, Dheeraj Kumar, ArvinderPal, Yash Sonker, Ritu Garg

Department of Computer Engineering, National Institute of
Technology, Kurukshetra, India
`gargginni01@gmail.com, kdheeraj922@gmail.com,`
`arvinderpal10@gmail.com,`
`yashsonker1998@gmail.com,ritu.59@gmail.com`

**Abstract.** There are many challenges in the classification of hyper spectral images such as large dimensionality, scarcity of labeled data and spatial variability of spectral signatures. In this proposed method, we make a hybrid classifier (MLP-SVM) using multilayer perceptron (MLP) and support vector machine (SVM) which aimed to improve the various classification parameters such as accuracy, precision, recall, f-score and to predict the region without ground truth. In proposed method, outputs from the last hidden layer of the neural network become the input to the SVM, which finally classifies into various desired classes. In the present study, we worked on Indian Pines, U. Pavia and Salinas dataset with 16, 9, 16 classes and 200, 103 and 204 reflectance bands respectively, which is provided by AVIRIS and ROSIS sensor of NASA Jet propulsion laboratory. The proposed method significantly increases the accuracy on testing dataset to 93.22%, 96.87%, 93.81% as compare to 86.97%, 88.58%, 88.85% and 91.61%, 96.20%, 90.68% based on individual classifiers SVM and MLP on Indian Pines, U. Pavia and Salinas datasets respectively.

## 1 Introduction

Remote sensing alludes to the process of acquisition of valuable information about distant target i.e. a natural phenomenon or an object on earth surface, without establishing any physical contact with it. It plays an important role in various fields like mapping of used land, environmental study, forecasting the weather and weather reports, study including natural hazards and exploration of resource. Land Cover Classification is the one we are focused on in this current work. This in turn plays an important role in urban planning, agriculture precision, management of land resources and environmental protection. Land cover is the word used for grass, water bodies, barren land etc. The required information is gathered with the help of aircrafts or satellites that collect information from the earth's surface. For almost every classification method in remote sensing, there are two main aspects, which are considered important: feature extractor and the classifier. The discriminative feature vectors are formed by transformation of data represented in the spectral, spatial and /or temporal form.

The function of classifier is to map each feature to a certain types. The images provided by the satellites are in the form of various spectral bands that are characterised based on their wavelengths and bandwidth. There are various types of hyper spectral cameras which are used for collecting the useful data from earth surface, these cameras are specified for particular wavelengths and bandwidths. In this study, we are using hyper-spectral images captured by AVIRIS (Airborne Visible/ Infrared Imaging Spectrometer) sensor and ROSIS sensor of NASA Jet propulsion laboratory. This provides calibrated images in 224 and 103 spectral bands with the wavelength ranging between 0.4-0.45 μm and 0.43-0.86 μm respectively. The AVIRIS is used for identification, measurements and monitoring purposes of constituents of earth's surface and atmosphere based on the index of molecular absorption and particle scattering signature.

Images with single band can be represented as a greyscale image in which each geographical area is represented by a particular value of spectral band. In spite of spatial –temporal images, remote sensing data in other form such as polar metric SAR (Synthetic Aperture Radar), with their extracted features are used as input for the classification. Selection of Bands with suitable approaches aims to find the bands that are most relevant and hence they keep a track of Spectral Information of original Hyper spectral Image. Band selection can be characterised as supervised, semi- supervised and unsupervised on the basis of prior information available to us. In case of supervised band selection, Distance Metrics are used to find most discerning spectral bands. It needs the labelled data in order to give desired result. In case of Unsupervised band selection, most informative and distinctive spectral bands with the help of spectral band ranking or spectral band clustering may also be used. In Semi supervised band selection process the most discerning and informative spectral band are find with the help of hyper graph model, mutual information(MI) or spectral band clustering. In recent years, various learning models have been proposed for HSI spectral spatial classification that claims to give higher performances. Convolutional Neural Network can be taken as a good example for the same, which gives acceptable results without flattening of training samples.

## 2   Literature Review

In the beginning, the land cover/ land use studies were majorly focused on the physical aspects of the changes but later their focused has moved from simplicity to realism over the recent year. In 2001, Land cover change analysis was done at Gurur Ganga watershed in Uttaranchal [1]. In 2009 ref. [2] observed the conversion of crops under crops to grassland which he said is related with the swiftly changes in the social, economical, demographical conditions after 1989. In 2010, in paper [3], authors studied the land cover/land use in which they observed the increase in the area under built-up land and harvested land whereas significant reduction in the forest land and area under different water bodies. In ref. [4], authors studied the land cover changes because of the mining activities from year 2001 to 2010. The results of study revealed the noticeable decrease in Forest Land, Land under cultivation and area covered by water bodies while significant increase in wasteland and land under anthropogenic activities. Authors in [5], showed up-gradation in land reforms and its resettlement. They concluded study with the significant decrease in forest land due to Human activities have become the major concern as these are the reason for the deterioration of the water quality index. Ref. [6] gives correlation between land use and quality index of water aids in the identification of threats to quality of river water. In year 2017, authors [7] proposed a multilayer deep learning architecture which was used for the classification of multi-temporal multisource RS images. The proposed architecture was applied for the classification of crops using two datasets such as Sentinel-1A time series and Landsat-8 that claimed to produce accuracy more than 85%. In the same year, authors in [8] proposed functional connectivity-based classification using CNN architecture. Ref. [9], proposed Land cover and crop classification using multi-temporal sentinel-2 images based on crops phenol-logical cycle in which they used Sentinel-2 dataset. They used Random Forest classifier for extracting the best images for the multi-temporal analysis.

## 3  Material and Methods

**Dataset**

In proposed method, experiments are conducted on Indian Pines, U. Pavia and Salinas dataset with 16, 9, 16 classes and 200, 103 and 204 bands of reflectance respectively as features for the classification process. Training and Testing dataset is divided into 4:1 respectively 10. The values of the pixels are normalized in the range [0, 1] using the following formula:

$$y = \frac{x - mi}{max(x)} \qquad (1)$$

where **x** represents pixel value and y represents the normalized pixel value.
The detailed information of the dataset used is shown in Table 1.

Table 1. Represents the information about dataset used in present study [10].

|  | Indian Pines | U. Pavia | Salinas |
|---|---|---|---|
| Sensor | AVIRIS | ROSIS | AVIRIS |
| Place | Northwestern Indiana | Pavia, Northern Italy | Salinas Valley California |
| Frequency Band | 0.4-0.45 μm | 0.43-0.86 μm | 0.4-0.45 μm |
| Spatial Resolution | 20m | 1.3m | 20m |
| No. of Channels | 220 | 103 | 220 |
| No. of Classes | 16 | 9 | 16 |

**Support Vector Machine**

Support Vector Machines, commonly referred to as SVM is a class of machine learning algorithms that works by finding a (n-1)-dimensional hyper-plane that correctly classifies data points, provided we have n-dimensional feature. It finds a plane by maximizing distance between the data points of available classes with the help of support vectors. They employ the use of kernels that is a set of mathematical functions, which takes data as input, and transform it to output. Some examples are linear, non-linear, polynomial, radial basis function (rbf) and sigmoid.

Support vectors are those data points or samples that are close to hyper-plane responsible for deciding distance and orientation of these hyper-planes. While in case of logistic regression we map the output to [0, 1] range with the help of sigmoid function whereas in SVM the output is mapped to [-1, 1] range which provides large confidence in classifying data points that it has never seen. It can be used both for classification and regression problems.

Hinge loss function is used in finding optimal positioning and orientation of hyper-planes as follows:.

$$L(y) = max\ (0,\ 1\text{-}t^*y) \qquad (2)$$

where **y** = given label and t = predicted class score and L(**y**) is loss function and the goal of SVM is to minimize this loss function.

### Multi Layer Perceptron

Commonly referred to as MLP, it is basically a combination of different layers of perceptron stacked together i.e. the output of first layer is used as input for the second layer and so on. The output layer gives a class score, which can then be fed to some activation function depending on the required output. Hidden layers are present between input layer and output layer with defined neurons by user. Training in MLP occurs in two steps such as Feed-forward and Back-propagation. Feed-forward is basically the process of converting input into output. Based on the output the loss is calculated using the proper loss function and then the back-propagation step is used to update the weights and biases of the network with the help of gradient descent. This process goes on some finite number of epochs as specified by user and then the model can be used for testing purposes. The back-propagation is used to update the weights using gradient descent that relies on chain rule of differentiation for updating weights.

## 4 Proposed Method

In the proposed method, we worked on pixel wise classification with patch size one using hybrid classifier MLP-SVM. The method consist of three hidden layers of the neural network with neurons 500, 350, 250 respectively, then output of the final hidden layer become the input to the SVM, which further classify it into different classes. The method is little more complex but overall it improves various parameters such as accuracy, precision, recall and f-score when compared to SVM and MLP classifiers individually. The hybrid MLP-SVM is shown in Fig 1. We are getting more accurate results using hybrid classifier because last hidden layer of MLP contains more significant features for each training data, which is further classified more accurately by using SVM.

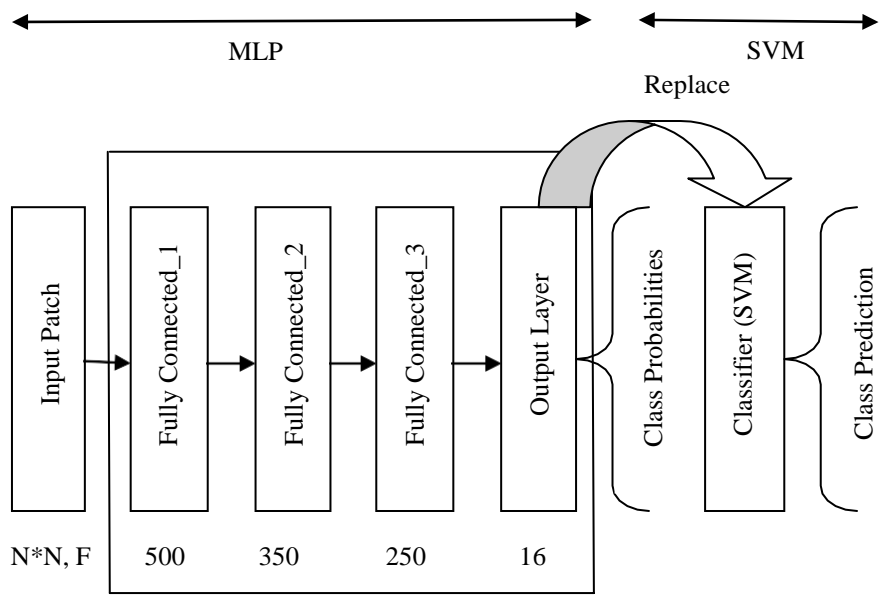

*N denotes input patch, Here N=1 and F is no. of Features

**Fig. 1** Represents the proposed hybrid classifier MLP-SVM.

Let's take an example to understand the working of the hybrid classifier, for instance take a Indian Pines dataset with dimensions 145*145*200 and 16 output classes, that is we have 21025 dataset instances each with F = 200 features (spectral Bands). Now, firstly we divide the dataset as 4:1 into training and testing dataset, followed we trained MLP with training dataset. Afterwards, we extracted weights matrix from hidden layers of dimensions 200*500 from first hidden layer, of dimensions 500*350 from second hidden layer, of dimensions 350*250 from final hidden layer. After all these we perform following operations as follow:

1. (21025*200) Matrix Multiply (200*500) gave 21025*500 matrix
2. RELU activation function on each value of matrix 21025*500
3. (21025*500) Matrix Multiply (500*350) gave 21025*350 matrix
4. RELU activation function on each value of matrix 21025*350
5. (21025*350) Matrix Multiply (350*250) gave 21025*250
6. Finally from dataset of size 21025, where each contains features 250 was divided into 4:1 for training and testing of SVM Classifier.
7. Finally, prediction of particular class was done by SVM Classifier.

## 5 Results and Discussion

The following metrics are used to evaluate the robustness of the proposed method.
**Overall accuracy:** It is calculated as the fraction of correctly classified labels over samples of all classes. Let **TP**, TN, FP, FN denote respectively, the number of true positive, true negative, false positive and false negative, then,

$$Accuracy = \frac{TP+TN}{TP+TN+FP+FN} \quad (3)$$

**Class- specific precision, Recall, F-score:** They can be calculated as the fraction of correctly classified labels over sample of specific class.

$$Precision = TP/(TP+FP) \quad (4)$$

$$Recall = TP/(TP+FN) \quad (5)$$

$$F\text{-}score = 2*TP/(2*TP + FP + FN) \quad (6)$$

Various parameters of proposed MLP-SVM method such as precision, recall, f-score and accuracy are compared with MLP-K-nearest neighbor (KNN), MLP-Random forest (RF), MLP-Decision Tree (DT), SVM, KNN, RF, DT, and MLP. Our results outperform the other methods in most of evaluation metrics used in present study.There is significant increase in precision, recall, f-score and accuracy metrics by 4.75%, 4.5%, 5.69% and 6.25% respectively on Indian Pines, when we compare our proposed method MLP-SVM with each of MLP and SVM. All the experiments are performed using Google Colaboratory. The detailed experimental computations on different datasets are shown in Table 2, Table 3 and Table 4.

**Table 2.** Represents the evaluation metrics based on hybrid classifier on Indian Pines.

|         | Precision% | Recall% | F-score% | Accuracy% |
|---------|------------|---------|----------|-----------|
| **MLP-SVM** | **92.25** | **93.75** | **92.88** | **93.22** |
| MLP-KNN | 88.06 | 88.13 | 87.75 | 90.83 |
| MLP-RF  | 93.13 | 91.81 | 92.31 | 91.41 |
| MLP-DT  | 86.31 | 86.75 | 86.31 | 86.24 |
| **SVM** | 87.50 | 89.25 | 87.19 | 86.97 |
| KNN     | 76.56 | 78.50 | 76.18 | 76.92 |
| RF      | 83.13 | 79.06 | 80.75 | 86.34 |
| DT      | 70.81 | 69.00 | 69.50 | 74.10 |
| **MLP** | 86.06 | 84.56 | 85.13 | 91.61 |

**Table 3.** Represents the evaluation metrics based on hybrid classifier on U. Palvia.

|         | Precision% | Recall% | F-score% | Accuracy% |
|---------|------------|---------|----------|-----------|
| **MLP-SVM** | **96.22** | **95.67** | **96.22** | **96.87** |
| MLP-KNN | 95.00 | 94.55 | 94.67 | 95.82 |
| MLP-RF  | 95.89 | 94.78 | 95.11 | 96.39 |
| MLP-DT  | 92.56 | 92.33 | 92.44 | 93.81 |
| **SVM** | 81.22 | 78.11 | 79.00 | 88.58 |
| KNN     | 91.66 | 89.56 | 90.44 | 91.22 |
| RF      | 93.00 | 90.44 | 91.56 | 92.85 |
| DT      | 87.00 | 87.33 | 87.33 | 88.95 |
| **MLP** | 95.11 | 95.22 | 95.22 | 96.20 |

**Table 4.** Represents the evaluation metrics based on hybrid classifier on Salinas.

|         | Precision% | Recall% | F-score% | Accuracy% |
|---------|------------|---------|----------|-----------|
| **MLP-SVM** | **97.19** | **97.31** | **97.25** | **93.81** |
| MLP-KNN | 96.19 | 96.25 | 96.25 | 92.23 |
| MLP-RF  | 97.63 | 97.56 | 97.69 | 94.65 |
| MLP-DT  | 96.44 | 96.63 | 96.44 | 92.34 |
| **SVM** | 93.86 | 93.00 | 93.00 | 88.85 |
| KNN     | 96.31 | 96.38 | 96.44 | 92.67 |
| RF      | 97.44 | 97.56 | 97.56 | 94.93 |
| DT      | 95.81 | 95.75 | 95.75 | 92.10 |
| **MLP** | 94.94 | 94.50 | 94.56 | 90.68 |

Bar graph is used to represent statistical data with rectangular bar with lengths of bar represents the value of corresponding evaluation metrics. The proposed method performs quite well on various evaluation metrics such as precision, recall, f-score and accuracy. The method outperform each of SVM and MLP, which shows its robustness as shown in Fig 2., Fig 3.and Fig 4., where vertical axis represents the experiment values and horizontal axis represents the various evaluation metrics. These graphs are used to increase the visualization of experimental results.

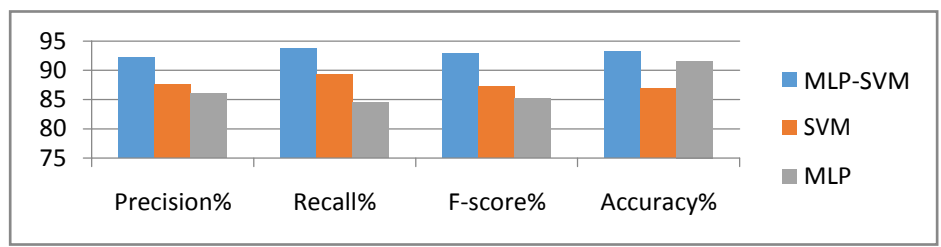

**Fig. 2** Represents the comparative study of proposed method on Indian Pines dataset.

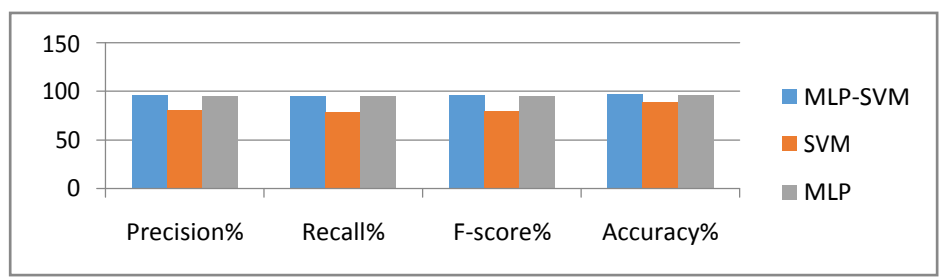

**Fig. 3** Represents the comparative study of proposed method on U. Palvia dataset.

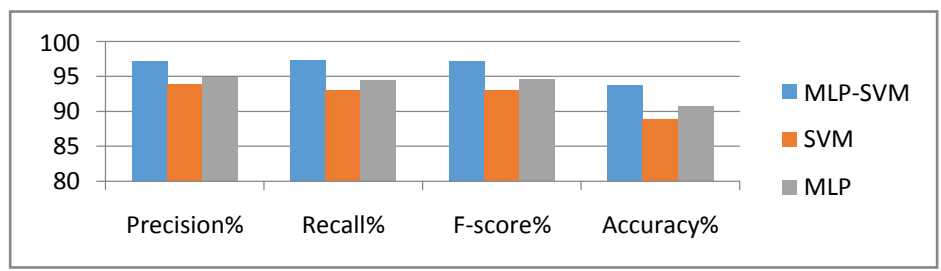

**Fig. 4** Represents the comparative study of proposed method on Salinas dataset.

Ground truth of Indian Pines, U. Pavia and Salinas dataset is shown in Fig 5. (a, c, e), which represents the actual labels of the different classes to be classified using supervised learning. Decoded output from our model for the different datasets is shown in Fig 5. (b, d, f). We also compute the region without ground truth in proposed method. The outputs labels are represented using the spectral library of python.

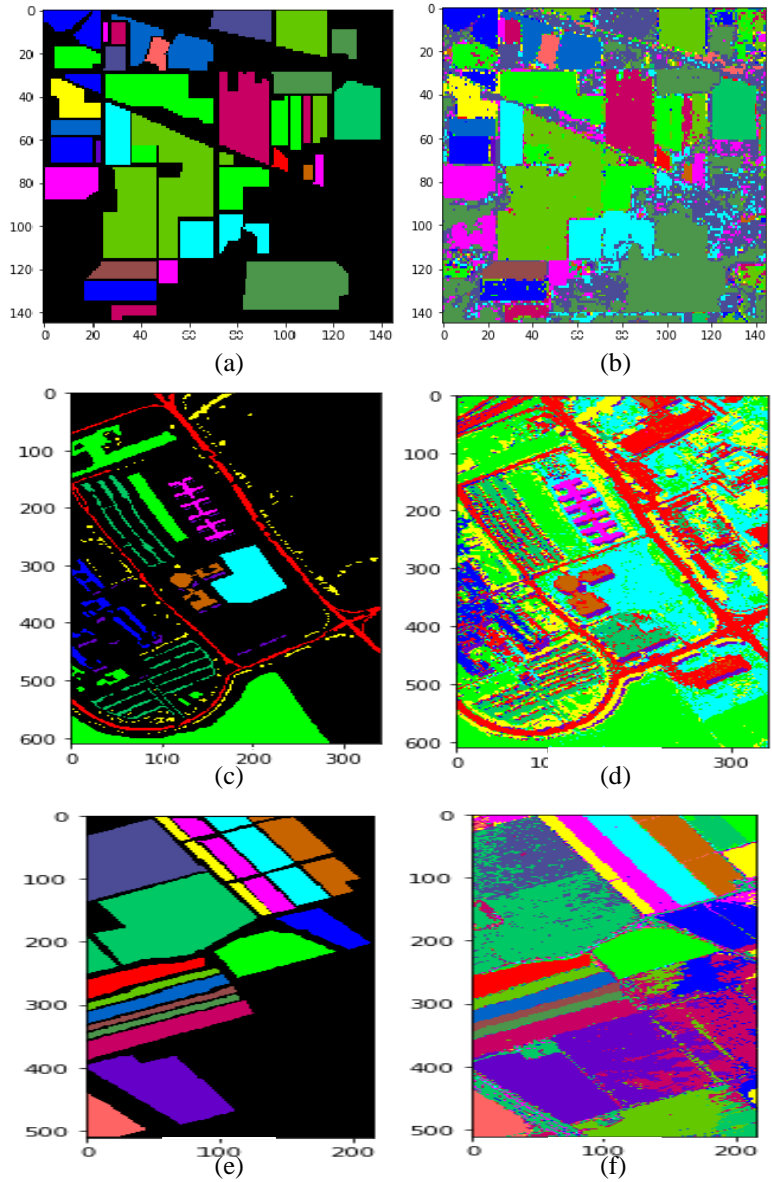

**Fig. 5** Represents the Indian Pines, U. Palvia and Salinas dataset (a), (c), (e) Ground Truth , (b), (d), (f) Output from proposed model with region without ground truth.

## 6      Conclusion

In the proposed model, we worked on hybridization of MLP-SVM. We used three hidden layers with neurons as 500, 350 and 250 respectively. Output of final hidden layer is used as input to the SVM for classification into different classes. We used patch size of one for spatial relationship with reflectance of different bands as features based on fraction of input to output flux of incident rays. All the experiments are conducted using Google Colaboratory. The proposed method outperforms MLP-KNN, MLP-RF, MLP-DT, SVM, KNN, RT, DT, and MLP in various evaluation metrics such as accuracy 93.22%, 96.87% and 93.81% on Indian Pines, U. Pavia and Salinas's dataset respectively. Overall, our proposed model is novel and robust. In future, it could be extended to other applications as well to check its robustness.